\documentclass{article}

    \usepackage[final,nonatbib]{neurips_2020}

\usepackage[utf8]{inputenc} % allow utf-8 input
\usepackage[T1]{fontenc}    % use 8-bit T1 fonts
\usepackage{hyperref}       % hyperlinks
\usepackage{url}            % simple URL typesetting
\usepackage{booktabs}       % professional-quality tables
\usepackage{amsfonts}       % blackboard math symbols
\usepackage{nicefrac}       % compact symbols for 1/2, etc.
\usepackage{microtype}      % microtypography
\usepackage{soul}           % strike-through (\st)
\usepackage{bm}             % boldface math
\usepackage{amsmath}        % for the use of \eqref
\usepackage{wrapfig,lipsum,booktabs}

\usepackage{graphicx}
\usepackage{multirow}

\usepackage[table]{xcolor}
\definecolor{Blue1}{RGB}{214, 235, 245}
\definecolor{Blue2}{RGB}{235, 245, 250}

\usepackage{color}
\usepackage{wrapfig}

\newcommand{\hightlightmodelname}[1]{{{{\underline{L}ook \underline{R}ead \underline{T}hink \underline{A}nswer}}}{#1}}
\newcommand{\modelabbrevname}[1]{{{{LRTA}}}{#1}}

\title{
\modelabbrevname{}: A Transparent Neural-Symbolic Reasoning Framework 
with Modular Supervision 
\\ for Visual Question Answering 
}

\author{Weixin Liang \\
  Stanford University \\
  \texttt{wxliang@stanford.edu} \\\And
  Feiyang Niu \\
  Amazon Alexa AI \\
  \texttt{nfeiyan@amazon} \\\And
  Aishwarya Reganti \\
  Amazon Alexa AI \\
  \texttt{areganti@amazon.com} \\\And
  Govind Thattai \\
  Amazon Alexa AI \\
  \texttt{thattg@amazon.com} \\\And
  Gokhan Tur \\
  Amazon Alexa AI \\
  \texttt{gokhatur@amazon.com}
}

\begin{document}

\maketitle

\begin{abstract}

The predominant approach to visual question answering (VQA) relies on encoding the image and question with a ``black-box'' neural encoder and decoding a single token as the answer like ``yes'' or ``no''. Despite this approach's strong quantitative results, it struggles to come up with intuitive, human-readable forms of justification for the prediction process. To address this insufficiency, we reformulate VQA as a full answer generation task, which requires the model to justify its predictions in natural language. We propose LRTA [Look, Read, Think, Answer], a transparent neural-symbolic reasoning framework for visual question answering that solves the problem step-by-step like humans and provides human-readable form of justification at each step. Specifically, LRTA learns to first convert an image into a scene graph and parse a question into multiple reasoning instructions. It then executes the reasoning instructions one at a time by traversing the scene graph using a recurrent neural-symbolic execution module. Finally, it generates a full answer to the given question with natural language justifications. Our experiments on GQA dataset show that LRTA outperforms the state-of-the-art model by a large margin (43.1\% v.s. 28.0\%) on the full answer generation task. We also create a perturbed GQA test set by removing linguistic cues (attributes and relations) in the questions for analyzing whether a model is having a smart guess with superficial data correlations. We show that LRTA makes a step towards truly understanding the question while the state-of-the-art model tends to learn superficial correlations from the training data.

\end{abstract}

\section{Introduction}
% \vspace{-0.35cm}
A long desired goal for AI systems is to play an important and collaborative role in our everyday lives~\cite{MOSS,DBLP:conf/acl/LiangZY20}. 
Currently, the predominant approach to visual question answering (VQA) relies on encoding the image and question with a black-box transformer encoder~\cite{lxmert,ViLBERT}. 
These works carry out complex computation behind the scenes but only yield a single token as prediction output (e.g., ``yes'', ``no''). Consequently, they struggle to provide an intuitive and human readable form of justification consistent with their predictions. 
In addition, recent study has further demonstrated some unsettling behaviours of those models: they tend to ignore important question terms~\cite{DBLP:conf/acl/MudrakartaTSD18}, look at wrong image regions~\cite{DBLP:conf/emnlp/DasAZPB16}, or undesirably adhere to superficial or even potentially misleading statistical associations~\cite{DBLP:conf/emnlp/AgrawalBP16}.

To address this insufficiency, we reformulate VQA as a full answer generation task rather than a classification one, i.e. a single token answer. The reformulated VQA task requires the model to generate a full answer with natural language justification. We find that the state-of-the-art model answers a significant portion of the questions correctly for the wrong reasons. 
To learn the correct problem solving process, 
We propose \modelabbrevname{} (\hightlightmodelname), a transparent neural-symbolic reasoning framework that solves the problem step-by-step mimicking humans.
    A human would first (1) \underline{l}ook at the image, (2) \underline{r}ead the question, (3) \underline{t}hink with multi-hop visual reasoning, 
    and finally (4) \underline{a}nswer the question. 
    Following this intuition, \modelabbrevname{} deploys four neural modules, each mimicking one problem solving step that humans would take:
    A scene graph generation module first converts an image into a scene graph; A semantic parsing module parses each question into multiple reasoning instructions; A neural execution module  interprets reason instructions one at a time by traversing the scene graph in a recurrent manner and; A natural language generation module generates a full answer containing natural language explanations. The four modules are connected 
    through hidden states rather than explicit outputs. 
    Therefore, the whole framework can be trained end-to-end, from pixels to answers.
    In addition, since \modelabbrevname{} also produces human-readable 
    output from individual modules during testing, we can easily
    locate the error by checking the modular output. 
    Our experiments on GQA dataset show that 
    \modelabbrevname{} outperforms the state-of-the-art model by a large margin 
    (43.1\% v.s. 28.0\%) on the full answer generation task. 
    Our perturbation analyses by removing relation linguistic cues from questions 
    confirm that 
    \modelabbrevname{} makes a step towards truly understanding the question rather than having a smart guess with superficial data correlations. 
    We discuss related work in Appendix A. To summarize, the main contributions of our paper are three-fold:
    \begin{itemize}
        \item We formulate VQA as a full answer generation problem (instead of short answer classification) to improve explainability and discourage superficial guess for answering the questions. 
        \item We propose \modelabbrevname{}, an end-to-end trainable, modular VQA framework facilitating explainability and enhanced error analysis 
        % via plug-and-play 
        as compared to contemporary black-box approaches. 
        \item We create a perturbed GQA test set that provides an efficient way to peak into a model's reasoning capability and validate our approach on the perturbed dataset. The dataset is available for future research - 
        \url{https://github.com/Aishwarya-NR/LRTA\_Perturbed\_Dataset}
        % \href{https://github.com/Aishwarya-NR/LRTA_Perturbed_Dataset}{\color{blue}{{https://github.com/Aishwarya-NR/LRTA\_Perturbed\_Dataset}}} 
    \end{itemize}

\begin{figure}[tb]
\centering
\includegraphics[width=0.80\textwidth, height=8cm]
{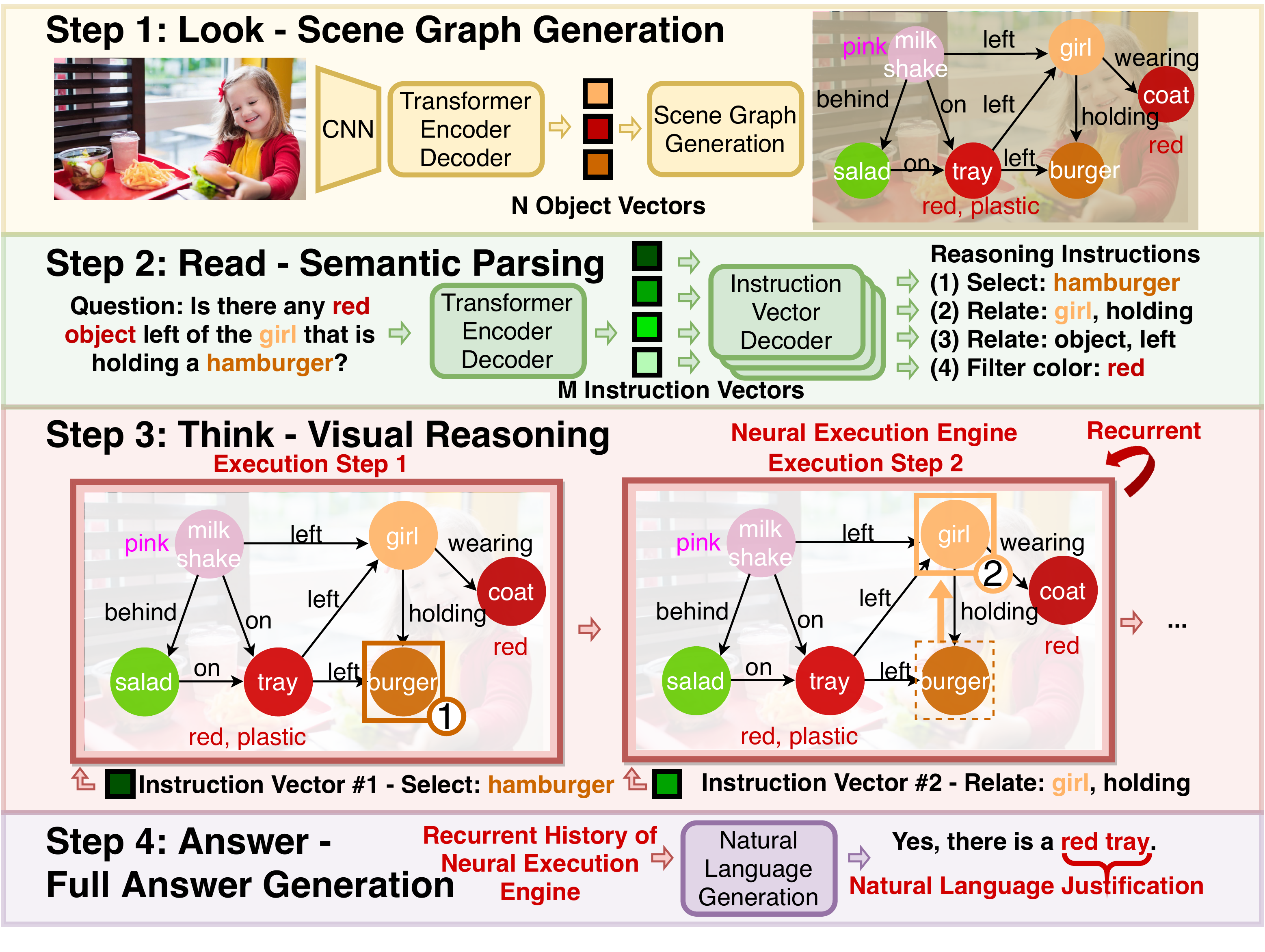} 
% \vspace{-4mm}
\caption{
\small
\modelabbrevname's four-step workflow 
(\underline{L}ook, \underline{R}ead, \underline{T}hink, \underline{A}nswer):  
(1) Convert the image into a scene graph (2) Parse the question into multiple reasoning instructions 
(3) Executes each instruction step-by-step using a recurrent neural execution engine 
(4) Generates the full answer with natural language justification. 
We connect four modules through hidden states rather than symbolic outputs so that the framework is end-to-end trainable.  
}
% \vspace{-6mm}
\label{fig:main}
\end{figure}

% \vspace{-0.45cm}
\section{\modelabbrevname{}: Look, Read, Think and Answer}
% \vspace{-0.25cm}
\paragraph{Look: Scene Graph Generation}
Given an image $I$, 
its corresponding scene graph represents the objects in the image (e.g., girl, hamburger) as nodes and the objects' pairwise relationships (e.g., holding) as edges. 
The first step of scene graph generation is object detection. 
We use DETR~\cite{detr} as the object detection backbone since it removes the need for for hand-designed components like non-maximum suppression. 
DETR~\cite{detr} feeds the image feature from ResNet50~\cite{resnet}
into a non-autoregressive transformer model, yielding an order-less set of $N$ object vectors $[\boldsymbol{o}_1, \boldsymbol{o}_2, \dots, \boldsymbol{o}_N]$, as in \eqref{eq:object_vector}.
Each object vector represents one detected object in the image. 
Then, for each object vector, 
DETR uses an object vector decoder (feed-forward network) to predict the corresponding object class (e.g., girl), and the bounding box in a multi-task manner. 
Since the set prediction of $N$ object vectors is order-less, DETR calculates the set prediction loss 
by first computing an optimal matching between predicted and ground truth objects, and then sum the loss from each object vector. 
$N$ is fixed to $100$ 
and DETR creates a special class label ``{\tt no object}'', to represent that the object vector does not represent any object in the image.
\begin{equation}
    [\boldsymbol{o}_1, \boldsymbol{o}_2, \dots, \boldsymbol{o}_N] = \mbox{DETR}(I)
\label{eq:object_vector}
\end{equation}
The object detection backbone learns object classes and bounding boxes, 
but does not learn object attributes, 
and the objects' pairwise relationships. 
We augment the object vector decoder with an additional object attributes predictor. 
For each attribute meta-concept (e.g., color), we create a classifier to predict the possible attribute values (e.g., red, pink). 
To predict the relationships, we consider all $N(N-1)$ possible pairs of object vectors,$[\boldsymbol{e}_1, \boldsymbol{e}_2, \dots, \boldsymbol{e}_{N(N-1)}]$. 
The relation encoder transforms each object vector pair to an edge vector through feed-forward and normalization layers as in \eqref{eq:edge_vector}. 
We then feed each edge vector to the relation decoder to classify its relationship label.
Both object attributes and inter-object relationships are supervised in a multi-task manner.
To handle the object vector pair that does not have any relationship, 
we use the ``{\tt no relation}'' relationship label. 
We construct the scene graph represented by $N$ object vectors and $N(N-1)$ edge vectors instead of the symbolic outputs, and pass it to downstream modules.  
% \textcolor{red}{\st{In practice, to reduce computation and memory overhead, 
% we only keep the object vectors with object detection confidence higher than $0.2$, and up to $500$ edge vectors with the most confident relationship predictions.}}
%
\begin{equation}
    \boldsymbol{e}_{i, j} = \mbox{LayerNorm}(\mbox{FeedForward}(\boldsymbol{o}_i \oplus \boldsymbol{o}_j))
\label{eq:edge_vector}
\end{equation}
%
% \paragraph{Graph Pruning} 
% \noindent \textbf{Graph Pruning} 
% The aforementioned scene graph imposes high computation and memory overhead for downstream modules. 
% Therefore, we only keep the object vectors with object detection confidence higher than $0.2$, and up to $500$ edge vectors with the most confident relationship predictions. 

% \vspace{-0.35cm}
\paragraph{Read: Semantic Parsing}
The semantic parser works as a ``compiler'' that 
translates the question tokens ($q_1, q_2, \dots, q_Q$) into an neural executable program, which consists of multiple instruction vectors. 
We adopt a hierarchical sequence generation design: 
a transformer model~\cite{transformer} first parses the question into a sequence of $M$ instruction vectors, $[\boldsymbol{i}_1, \boldsymbol{i}_2, \dots, \boldsymbol{i}_M]$. 
The $i^{th}$ instruction vector will correspond exactly to the $i^{th}$ execution step in the neural execution engine. 
To enable human to understand the semantics of the instruction vectors, 
we further translate each instruction vector to human-readable text using a transformer-based instruction vector decoder. 
We pass the $M$ instruction vectors rather than the human-readable text to the neural execution module. 
\begin{equation}
    [\boldsymbol{i}_1, \boldsymbol{i}_2, \dots, \boldsymbol{i}_M] = \mbox{Transformer}(q_1, q_2, \dots, q_Q)
\label{eq:instruction_vector}
\end{equation}
%

% \vspace{-0.35cm}
\paragraph{Think: Visual Reasoning with Neural Execution Engine} 
The neural execution engine works in a recurrent manner: 
At the $m^{th}$ time step, 
the neural execution engine takes the $m^{th}$ instruction vector ($\boldsymbol{i}_m$) and 
outputs the scene graph traversal result. 
Similar to recurrent neural networks, 
a history vector that summarizes the graph traversal states of all nodes in the current time-step would be passed to the next time-step. 
The neural execution engine operates with graph neural network. 
Graph neural network generalizes the convolution operator to graphs using the 
neighborhood aggregation scheme~\cite{graphMetaNetwork,howPowerfulGNN}. 
The key intuition is that each node aggregates feature vectors of its immediate neighbors to compute its new feature vector as the input for the following neural layers. 
Specifically, at $m^th$ time step given a node as the central node, we first obtain the feature vector of each neighbor ($\boldsymbol{f}_k^m$) through a feed-forward network with the following inputs: the object vector of the neighbor ($\boldsymbol{o}_k$) in the scene graph, the edge vector between the neighbor node and the central node ($\boldsymbol{e}_{k, \mbox{central}}$) in the scene graph, the $(m-1)^{th}$ history vector ($\boldsymbol{h}_{m - 1}$), and the $m^{th}$ instruction vector ($\boldsymbol{i}_m$).
\begin{equation}
    \boldsymbol{f}_k^m = \mbox{FeedForward}(\boldsymbol{o}_k \oplus \boldsymbol{e}_{k, \mbox{central}} \oplus \boldsymbol{h}_{m - 1} \oplus \boldsymbol{i}_m)
\label{eq:feature_vector}
\end{equation}
We then average each neighbor's feature vector as the context vector of the central node ($\boldsymbol{c_{\mbox{central}}^m}$).
\begin{equation}
    \boldsymbol{c}_{\mbox{central}}^m = \frac{1}{K} \sum_{k = 1}^K \boldsymbol{f}_k^m
\label{eq:context_vector}
\end{equation}
Next, we perform node classification for the central node, where an ``{\tt 1}'' means that the corresponding node should be traversed at the $m^{th}$ time step and ``{\tt 0}'' otherwise. 
The inputs of the node classifier are: the object vector of the central node in the scene graph, the context vector of the central node, and the $m^{th}$ instruction vector. 
\begin{equation}
    s_{\mbox{central}}^m = \mbox{Softmax}(\mbox{FeedForward}(\boldsymbol{o}_{\mbox{central}} \oplus \boldsymbol{c}_{\mbox{central}}^m \oplus \boldsymbol{i}_m))
\label{eq:traversal_classifier}
\end{equation}
where $s_{\mbox{central}}^m$ is the classification confidence score of central node at $m^{th}$ time step.
The node classification results of all nodes constitute a bitmap as the scene graph traversal result. 
We calculate the weighted average of all object vectors as the history vector ($\boldsymbol{h}_m$), where the weight is each node's classification confidence score.
\begin{equation}
    \boldsymbol{h}_m = \sum_i^N s_i^m \cdot \boldsymbol{o}_i
\label{eq:history_vector}
\end{equation}
% We implement our model using PyTorch Geometric~\cite{pytorchGeometric}. 

% \vspace{-0.35cm}
\paragraph{Answer: Full Answer Generation}
VQA is commonly formulated as a classification problem where 
the model learns to answers with only one token (e.g., ``yes'' or ``no''). 
We advocate to formulate VQA as a natural language generation problem, 
where the model learns to answer the question in a full sentence with justifications. 
To do this, \modelabbrevname{} deploys a transformer model that takes in the neural execution's history vectors from all time-steps, and generates the full answer tokens ($a_1, a_2, \dots, a_A$). 
\begin{equation}
    (a_1, a_2, \dots, a_A) = \mbox{Transformer}(\boldsymbol{h}_1 \oplus \boldsymbol{h}_2 \oplus \dots \oplus \boldsymbol{h}_M)
\end{equation}
%

% \vspace{-0.35cm}
\paragraph{End-to-End Training: From Pixels to Answers}
    We connect four modules through hidden states rather than symbolic outputs~\cite{MOSS}. 
    Therefore, the whole framework could be trained in an end-to-end manner, from pixels to answers. 
    The training loss is simply the sum of losses from all four modules. 
    Each neural module receives supervision not only from the module's own loss, but also from the gradient signals back-propagated by downstream modules. 
    We start from the pre-trained weights of DETR for the object detection backbone and 
    all other neural modules are randomly initialized.

\section{Experiments}
% \vspace{-0.25cm}
\paragraph{Setup} 
We evaluate \modelabbrevname{} on the GQA dataset~\cite{GQA}, which contains 1.5M questions over 110K images. 
To the best of our knowledge, 
\modelabbrevname{} is the first full answer generation model on GQA~\cite{GQA}. 
We use the standard dataset split. 
During training, we use the ground truth for scene graphs, reasoning instructions, scene graph traversal results for each step, and full answers. During testing, we only use images and questions. 
We add transformer decoder to the state-of-the-art short answer model LXMERT~\cite{lxmert} as a full answer generation baseline. 
We report accuracy on both short answers and full answers for both LXMERT and \modelabbrevname{}. 
Full answers are evaluated with string match accuracy since the full answers follows pre-defined templates. 
We delay improving the metric as future work. 
% The validation split provides full annotations nand the testdev split provides incomplete annotations. We did not consider the test set since the annotations are not publicly available and the test server does not support full answer evaluation. 

\begin{table*}
\centering
% \begin{minipage}{.30\linewidth}
\begin{minipage}{.30\linewidth}
\vspace*{0.128em}

\centering
% \scriptsize 
\small 
\tabcolsep=0.05cm
\begin{tabular}{rccccccc}
\cmidrule[\heavyrulewidth]{1-3}
\textbf{Model} & \textbf{Full Acc} & \textbf{Short Acc} \\[0.2em]
\cmidrule{1-3}
Prior~\cite{GQA} & - & 28.93\% \\[0.2em]
Human~\cite{GQA} & - & 89.30\% \\[0.2em]
Bottom-up~\cite{bottomUp} & - & 49.74\% \\[0.2em]
MAC~\cite{MAC} & - & 54.06\% \\[0.2em]
LXMERT~\cite{lxmert} & 28.00\% & \bf 56.20\% \\[0.2em]
\bf \modelabbrevname{} & \bf 43.10\% & \bf 54.48\% \\[0.2em]
\cmidrule[\heavyrulewidth]{1-3}
\end{tabular}
\caption{End-to-end training \newline experiment on testdev set}
\label{tab:testdev}
\end{minipage}%
\hspace{2em}
\begin{minipage}{.60\linewidth}
\vspace*{0.128em}
\centering
% \scriptsize 
\small 
\tabcolsep=0.05cm
\begin{tabular}{lccccccc}
\cmidrule[\heavyrulewidth]{1-3}
\textbf{Model} & \textbf{Full Acc} & \textbf{Short Acc} \\[0.2em]
\cmidrule{1-3}
\multicolumn{3}{l}{\modelabbrevname{} trained w/ visual oracle} \\[0.2em]
\hspace{3mm}Evaluated w/o attributes & 67.79\% & 78.21\% \\[0.2em]
\hspace{3mm}Evaluated w/o relations & 67.95\% & 75.47\% \\[0.2em]
\hspace{3mm}Evaluated w/o attributes \& relations & 50.15\% & 61.15\% \\[0.2em]
\hspace{3mm}Evaluated w/ visual oracle & \bf 85.99\% & \bf 93.10\% \\[0.2em]
\cmidrule{1-3}
\multicolumn{3}{l}{\modelabbrevname{} trained w/ reading oracle} \\[0.2em]
\hspace{3mm}Evaluated w/ reading oracle & 55.45\% & 64.36\%  \\[0.2em]
\cmidrule[\heavyrulewidth]{1-3}
\end{tabular}
\caption{Validation study on valid set}
\label{tab:valid}
\end{minipage}%
\end{table*}

% \vspace{-0.30cm}
\paragraph{Design Validation with Ground Truth Scene Graph} 
Since \modelabbrevname{} deviates from the predominant black-box encoder approach a lot, 
we first validate the design of \modelabbrevname{} by using a visual oracle for step 1 (ground truth scene graphs). 
As shown in Table~\ref{tab:valid}, \modelabbrevname{} with visual oracle achieves a surprisingly high accuracy on both short answers (93.1\%) and full answers (85.99\%) on the validation set. This shows the great potential and expressivity of \modelabbrevname{} for visual question answering. 
In addition, if we remove the attributes or the relations in the test data, the performance drops a lot. This shows that scene graph generation beyond object detection is a crucial step and thus we call for more attention to scene graphs for the visual question answering community.

% \vspace{-0.30cm}
\paragraph{End-to-End Training Experiments} Next we train the model end-to-end, from pixels to answers. As shown in Table~\ref{tab:testdev}, \modelabbrevname{} significantly outperforms LXMERT in full answer generation (43\% v.s. 28\%) and achieves comparable accuracy on short answers (54.48\% v.s. 56.2\%). Next, we conduct perturbation study to show that the performance of LXMERT comes more from superficial data correlations while \modelabbrevname{} makes a step towards truly understanding the question.

\begin{wraptable}{r}{7.0cm}
% \begin{wraptable}{r}{5.5cm}
% \scriptsize
\small 
\begin{tabular}{lccccccc}
\hline
\textbf{Model} & \textbf{Short Acc Drop} (from $\rightarrow$ to) \\[0.2em]
\hline
\multicolumn{2}{l}{VB \& PRPN masked} \\[0.2em]
\hspace{3mm}LXMERT~\cite{lxmert} & 19.43\% (56.20\% $\rightarrow$ 36.77\%) \\[0.2em]
\hspace{3mm}\modelabbrevname{} & \textbf{26.20\%} (54.48\% $\rightarrow$ 28.28\%) \\[0.2em]
\hline
\multicolumn{2}{l}{Attributes masked} \\[0.2em]
\hspace{3mm}LXMERT~\cite{lxmert} & 9.41\% (56.20\% $\rightarrow$ 46.79\%) \\[0.2em]
\hspace{3mm}\modelabbrevname{} & \textbf{21.03\%} (54.48\% $\rightarrow$ 33.45\%) \\[0.2em]
\hline
\end{tabular}
\caption{Perturbation analysis on testdev set. The larger drop the better.}
\label{table3}
\end{wraptable}

% \vspace{-0.30cm}
% \paragraph{Adversarial Experiments: Perturbation Studies on testdev Set}
\paragraph{Perturbed GQA Dataset and Additional Analysis}
Finally, in order to probe whether a model has effectively leveraged linguistic cues, we design a perturbation study by systematically removing the cues such as attributes and relationships from the questions and evaluate if the model's performance changes significantly. Specifically, the better a model understands the language cues, the more drop we expect the model's performance on the cues stripped questions. We use a comprehensive list of attributes obtained by~\cite{metaModule} and mask them using a predefined mask token. For effectively masking relationships, we use Spacy POS-Tagger~\cite{spacy2} and mask \textit{verbs} (VB) and \textit{prepositions} (PRPN) from the question. The results are reported in Table~\ref{table3}. We evaluate LXMERT and \modelabbrevname{} for short answer accuracy on the testdev for consistency (Results on this analysis for the public valid set are reported in the Appendix). We can deduce from the results that \modelabbrevname{} results drop more significantly than LXMERT in both the masking scenarios. For relationships, we see that \modelabbrevname{}  performance drops by \textbf{26.20\%} as compared to 19.43\% drop in LXMERT, while for attributes, the margin is more significant at \textbf{21.03\%} and 9.41\% respectively, thus providing us a strong convergent evidence for our hypothesis that \modelabbrevname{} truly takes a leap forward while trying to systematically understand the question and its components rather than using peripheral correlations.

% \vspace{-0.35cm}
\section{Conclusion}
% \vspace{-0.25cm}
We present \modelabbrevname{}, a transparent neural-symbolic reasoning framework for visual question answering, that incorporates [look, read, think and answer] steps to provide a human-readable form of justification at each step. The modular design of our methodology enables the whole framework to be trainable end-to-end. Our experiments on GQA dataset show that \modelabbrevname{} achieves high accuracy on full answer generation task, outperforming the state-of-the-art LXMERT results by a noticeable 15\% absolute margin. In addition, \modelabbrevname{} performance drops significantly more than LXMERT, when object attributes and relationships are masked, hence indicating that \modelabbrevname{} makes a step forward, towards truly understanding the question, rather than making a smart guess based on superficial data correlations. In the validation study, we have shown that when provided with an oracle scene graph, \modelabbrevname{} is able to achieve a high accuracy on both short answers (93.1\%) and full answers (85.99\%), nearing the theoretical bound 96\% on short answers~\cite{DBLP:journals/corr/abs-2006-11524}. These observations indicate that better scene graph prediction methods offer a great potential in further improving \modelabbrevname{} performance on both short-answer and full-answer tasks.

\section*{Acknowledgement}
We would like to thank Robinson Piramuthu, Dilek Hakkani-Tur, Arindam Mandal, Yanbang Wang and the anonymous reviewers for their insightful feedback and discussions that have notably shaped this work.

\bibliographystyle{abbrv}
\bibliography{main}

\appendix
\newpage

\section*{Appendix A: Related Work}

\paragraph{Explainable VQA} 
Existing black-box visual question answering models attempt to directly map inputs to outputs using black-box architectures without explicitly modeling the underlying reasoning processes. 
To mitigate the black-box nature of these models, 
several model interpretation techniques have been developed to improve model transparency and explainability~\cite{DBLP:conf/emnlp/DasAZPB16,DBLP:conf/aaai/GhorbaniAZ19,DBLP:conf/eccv/LiTJCL18,DBLP:conf/emnlp/LiFYML18,liang-etal-2020-alice,ijcv}. 
One of the most popular approaches is attention map visualization, which highlights the important image regions for answering the question. 
However, recent study shows that 
the visualized attention regions correlate poorly with humans~\cite{DBLP:conf/emnlp/DasAZPB16,DBLP:conf/aaai/GhorbaniAZ19}. 
Another line of research propose to generate natural language justifications~\cite{DBLP:conf/eccv/LiTJCL18,DBLP:conf/emnlp/LiFYML18} along with the short answer. 
Similar to the GQA dataset~\cite{GQA}, 
the FSVQA dataset~\cite{fsvqa} provides full answer annotations for VQA, but the full answer generation task remains unexplored.  
To the best of our knowledge, 
\modelabbrevname{} is the first full answer generation framework for GQA~\cite{GQA}, and possibly for the visual quesiton answering task. 
However, this approach still do not reveal the model's step-by-step problem solving process. 
This approach makes a step towards more explainable VQA models, but still does not reveal the internal problem solving process of the model.

\paragraph{Neural Module Networks} 
Another active line of research on visual question answering 
explores neural module networks (NMN)~\cite{jacob,jacob2,jacob3,DBLP:conf/iccv/JohnsonHMHFZG17,DBLP:conf/eccv/HuADS18,DBLP:conf/cvpr/ShiZL19,DBLP:conf/icml/VedantamDLRBP19}, which composes the model's neural architecture on the fly based on the given question.  
Instead of training a model with static neural architecture, 
they hand-defined a set of small neural networks (i.e., neural modules), each dedicated for a specific kind of logical operation. 
Given a question, 
NMN first uses a semantic parser to parse the question into a series of logical operations (similar to our reasoning instructions). 
Then, given a sereis of logical operations, NMN dynamically layouts the small neural networks. 
For example, to answer ``What color is the metal cube?'', NMN dynamically composes four modules: 
(1) a module that finds things made of metal, (2) a module that localizes cubes, (3) a module that determines the color of objects. 
However, NMN models are challenging to optimize by its nature. Therefore, 
its success is mostly restricted to the
synthetic CLEVR dataset~\cite{DBLP:conf/iccv/JohnsonHMHFZG17} and how to extend NMN to real-world datasets is still an open research problem~\cite{metaModule}. 
Different from NMNs, our framework is conceptually simple and could be easily trained in an end-to-end manner, from pixels to answers.

\begin{wrapfigure}{r}{0.5\textwidth}
\centering
\includegraphics[width=0.5\textwidth]
{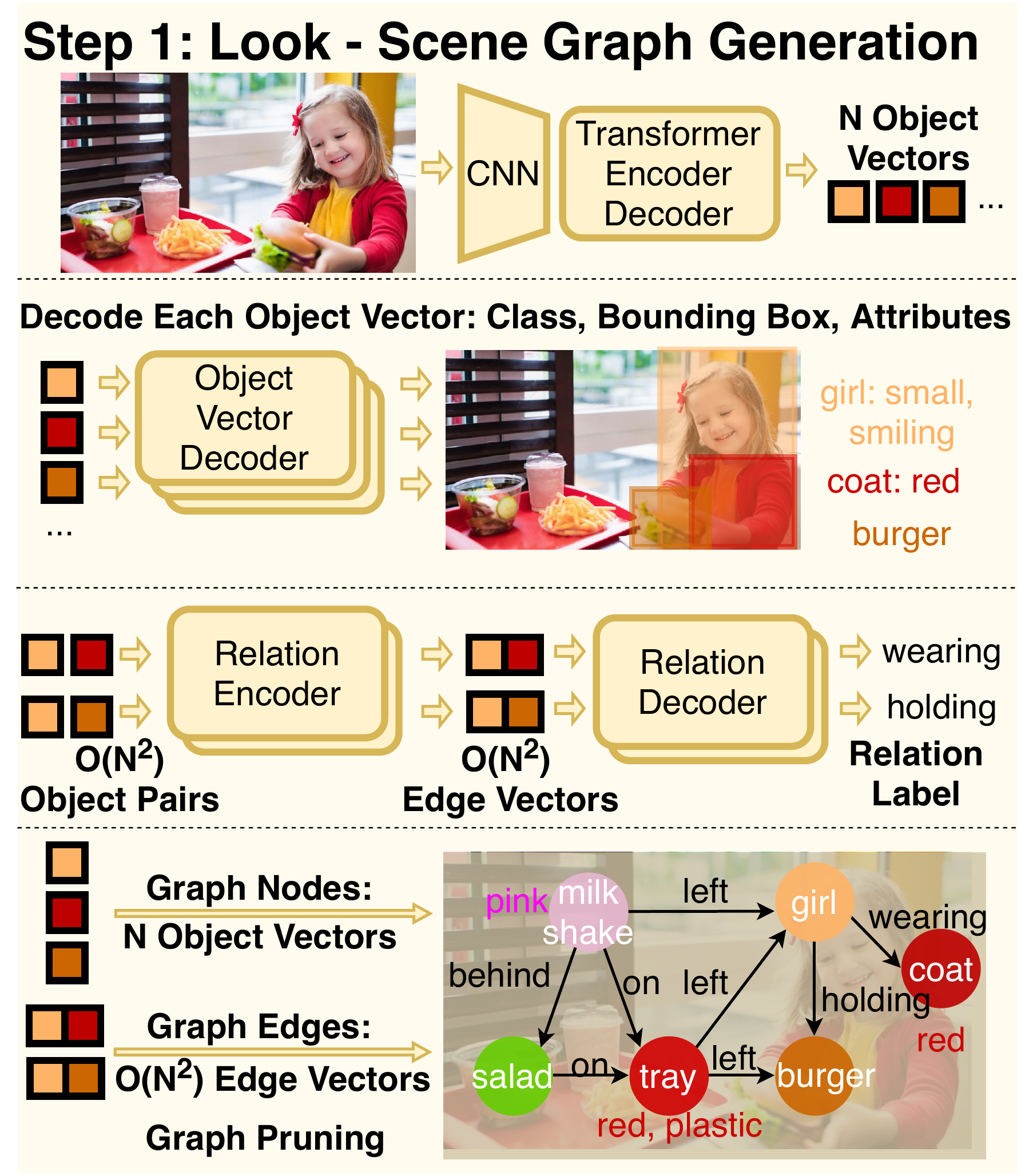}
% \vspace{-9mm}
% \vspace{-4mm}
\caption{
\small
\modelabbrevname's Scene Graph Generation Workflow.
}
\vspace{-6mm}
% \label{fig:mainAppendix}
\end{wrapfigure}

% \paragraph{Graph-based VQA} 
Our work is also related to the sporadic attempts in scene graph based VQA. 
Hudson et al.~\cite{neuralStateMachine} propose neural state machines that simulates the computation of an automaton on probabilistic scene graphs.  ~\cite{lcgn,DBLP:conf/iccv/LiGCL19} models the interaction between objects using graph attention mechanism. 
Different from their work, our work is end-to-end trainable, from pixels to answers, 
and transparently provides the execution result of each step.

\paragraph{Scene Graph Generation} Scene graph generation (SGG)~\cite{DBLP:conf/cvpr/XuZCF17} is a visual detection task that aims to predict objects and their relations from an image. In recent years, it has drawn increasing attentions that greatly advance the interface of vision and language. By extracting the concepts and contextual relations from pixels, scene graph provides an intuitive high-level summary of a raw image and facilitates down-stream reasoning tasks such as image captioning~\cite{DBLP:conf/iccv/GuJCZYW19,DBLP:conf/cvpr/YangTZC19,DBLP:conf/eccv/YaoPLM18}, VQA~\cite{DBLP:conf/cvpr/TeneyLH17,GQA,neuralStateMachine,DBLP:conf/bmvc/ZhangCX19} or image retrieval~\cite{DBLP:conf/cvpr/JohnsonKSLSBL15}. Previous works~\cite{DBLP:conf/cvpr/XuZCF17,neuralStateMachine,DBLP:conf/cvpr/ZellersYTC18,DBLP:conf/eccv/YangLLBP18,DBLP:conf/cvpr/ChenYCL19} have predominantly relied on Faster R-CNN based detectors that typically generate a potentially large set of bounding box proposals whose contextualized representation is then fed through a subsequent sequence to sequence network (e.g. LSTM~\cite{DBLP:conf/cvpr/ZellersYTC18} or Transformers~\cite{DBLP:journals/corr/abs-2004-06193}) to predict object labels and their relations. A key disadvantage of those detectors is that they normally need many hand-designed components like a non-maximum suppression procedure or anchor generation. In that regard, we adopted DETR~\cite{detr}, a recently proposed method that streamlines the detection process and makes our whole pipeline end-to-end trainable. Despite the growing research interest, SGG remains as a challenging task largely due to the training bias, e.g. {\tt <human, on, beach>} appears more frequently than {\tt <human, lay on, beach>}. As such, dummy models predicting solely based on frequency is embarrassingly not far from the state-of-the-art as reported in~\cite{DBLP:conf/cvpr/TangNHSZ20,DBLP:journals/corr/abs-2005-08230}. An unbiased SGG method~\cite{DBLP:conf/cvpr/TangNHSZ20} was recently proposed that sheds some promising light on the data bias issue. Thanks to the modular design of \modelabbrevname{}, we can easily incorporate such unbiased SGG into our pipeline.

\section*{Appendix B: Additional Experiments}

\begin{wraptable}{r}{7.0cm}
% \begin{wraptable}{r}{5.5cm}
% \scriptsize
\small 
\begin{tabular}{lccccccc}
\hline
\scriptsize \textbf{Model} & \textbf{Short Acc Drop} (from $\rightarrow$ to) \\[0.2em]
\hline
\multicolumn{2}{l}{VB \& PRPN masked} \\[0.2em]
\hspace{3mm}LXMERT~\cite{lxmert} & 4.40\% (64.30\% $\rightarrow$ 59.90\%) \\[0.2em]
\hspace{3mm}\modelabbrevname{} & \textbf{16.67\%} (62.79\% $\rightarrow$ 46.12\%) \\[0.2em]
\hline
\multicolumn{2}{l}{Attributes masked} \\[0.2em]
\hspace{3mm}LXMERT~\cite{lxmert} & 9.24\% (64.30\% $\rightarrow$ 55.06\%) \\[0.2em]
\hspace{3mm}\modelabbrevname{} & \textbf{16.57\%} (62.79\% $\rightarrow$ 46.22\%) \\[0.2em]
\hline
\end{tabular}
\caption{Perturbation analysis on valid set. The larger drop the better. }
\label{table4}
\end{wraptable}

\paragraph{Additional Adversarial experiments: Perturbation studies on valid-set}
The below table reports the perturbation study results on the valid dataset, we notice a similar pattern to the test-dev set. \modelabbrevname{}  has a higher drop in performance when the attributes/relationships are masked as compared to LXMERT~\cite{lxmert}.

\end{document}